# Detecting Intentions of Vulnerable Road Users Based on Collective Intelligence

Maarten Bieshaar, Günther Reitberger, Stefan Zernetsch,
Prof. Dr. Bernhard Sick, Dr. Erich Fuchs,
Prof. Dr.-Ing. Konrad Doll

University of Kassel
Wilhelmshöher Allee 71-73, 34121 Kassel,
0561/804-6056, {mbieshaar, bsick}@uni-kassel.de

University of Passau
Innstraße 41, 94032 Passau,
0851/509-3104, {reitberg, fuchse}@forwiss.uni-passau.de,

University of Applied Sciences Aschaffenburg
Würzburger Straße 45, 63743 Aschaffenburg,
06021/4206-871, {stefan.zernetsch, konrad.doll}@h-ab.de

**Abstract**

Vulnerable road users (VRUs, i.e. cyclists and pedestrians) will play an important role in future traffic. To avoid accidents and achieve a highly efficient traffic flow, it is important to detect VRUs and to predict their intentions. In this article a holistic approach for detecting intentions of VRUs by cooperative methods is presented. The intention detection consists of basic movement primitive prediction, e.g. standing, moving, turning, and a forecast of the future trajectory. Vehicles equipped with sensors, data processing systems and communication abilities, referred to as intelligent vehicles, acquire and maintain a local model of their surrounding traffic environment, e.g. crossing cyclists.

Heterogeneous, open sets of agents (cooperating and interacting vehicles, infrastructure, e.g. cameras and laser scanners, and VRUs equipped with smart devices and body-worn sensors) exchange information forming a multi-modal sensor system with the goal to reliably and robustly detect VRUs and their intentions under consideration of real time requirements and uncertainties. The resulting model allows to extend the perceptual horizon of the individual agent beyond their own sensory capabilities, enabling a longer forecast horizon. Concealments, implausibilities and inconsistencies are resolved by the collective intelligence of cooperating agents.

Novel techniques of signal processing and modelling in combination with analytical and learning based approaches of pattern and activity recognition are used for detection, as well as intention prediction of VRUs. Cooperation, by means of probabilistic sensor and knowledge fusion, takes place on the level of perception and intention recognition. Based on the requirements of the cooperative approach for the communication a new strategy for an ad hoc network is proposed. The evaluation is done using real data gathered with a research vehicle, a research intersection with public traffic and mobile devices.

# Motivation:

The World Health Organization reported in 2013 that by 2030 road traffic deaths will position themselves from the currently eighth up to the fifth leading cause of death unless urgent action is taken [1]. Cooperative, automated driving will be one of these actions in countries which presently rank first in road safety. 27% of 1.24 million victims worldwide are vulnerable road users (VRUs), e.g., pedestrians and cyclists [1]. The same percentage holds for VRUs in Germany with respect to all persons either killed or injured on the road in 2013 [2]. 101 873 accidents with injuries have been registered, 58% of the killed persons were pedestrians and 42% were cyclists. Most of the fatal accidents (69%) happened in urban areas. Many dangerous situations occur at intersections due to occlusions. An urban intersection, especially if equipped with traffic lights, presents natural accumulation points of pedestrians and bicyclists. Avoiding corresponding accidents is one objective of our current research [3]. Because of their significant accident rates, VRUs have to be indispensably considered also when reducing accidents and increasing traffic flow by cooperative, automated driving. Due to the ability of pedestrians and bicyclists to suddenly start a motion or to change the direction of motion, a dangerous situation may occur within some hundreds of milliseconds. Early and reliable detection of those movements, as well as the estimation of short-time VRU trajectories is a major challenge. To do it cooperatively in automated driving opens a new field of research.

In the proposed approach, we will address the problem of cooperative protection of VRUs. The collective intelligence of cars, complemented by information from infrastructure (where available, e.g., at intersections), and VRUs themselves (if equipped with intelligent devices such as smartphones) will be exploited to detect the intention of VRUs in a distributed way (e.g. Figure 1).

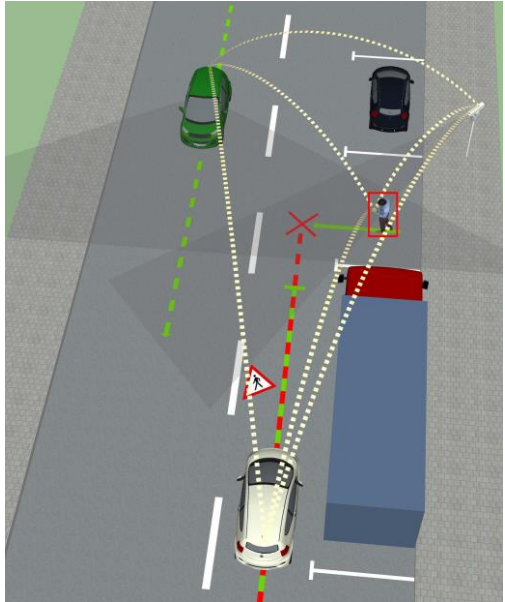

*Figure 1: Example scene with different agents.*

The approach does not only provide an essential component for future traffic automation, it also increases the safety of road users. Pedestrians and bicyclists are in the focus of our research. For the sake of brevity, we refer to these (and only these) as vulnerable road users. The term intention is generally used for a planned but not yet started or even finished action. We distinguish two aspects of intention detection. First, we see the motion of a VRU as a sequence of activities or basic movements, such as standing, walking, pedaling, stopping, starting, or bending. Our first aim is to detect movement transitions in order to forecast basic movements as early as possible. Second, we see the motion of a VRU as motion of certain body points (e.g. center of gravity, joints, or head) in the 3D space. Our second aim is to forecast trajectories of such points. An early knowledge about a movement transition can support trajectory forecast significantly. Both, basic movement forecasting and trajectory forecasting, are part of what we refer to as intention detection. They are based on a perception of VRUs in sensor data, e.g., video sequences, and characteristic attributes (features) extracted from those sensor data such as positions, velocities, accelerations, orientations, poses, view direction, etc. that describe the behavior of VRUs. A schematic representation of the intention detection model is depicted in Figure 2.

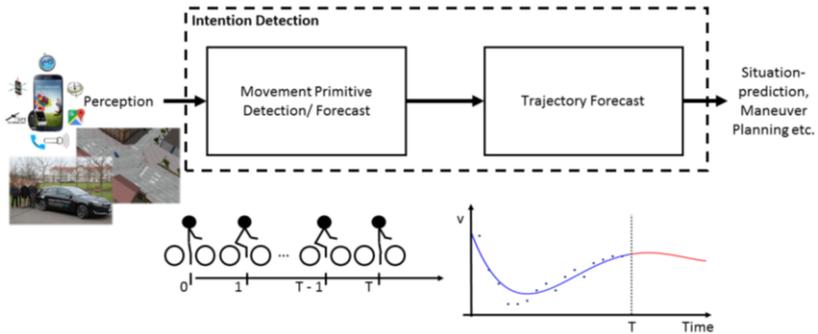

Figure 2: Schematic representation of the intention detection model.

Collective Intelligence (CI) is the joint intelligence or swarm intelligence of individuals or entities as a result of collective sensing (collective recording of data or knowledge), coordination of such individuals or entities, or even their cooperation. Individuals may be humans, software agents, intelligent technical systems, or mixtures of all. CI systems aim to reach either specific goals of the individuals, or superior goals of the overall system consisting of all individuals. Typically, CI systems exhibit properties such as adaptivity, interaction, self-organization, robustness, etc. The aim is cooperation of individuals, which are cars, infrastructure and VRUs themselves. In the following, these are termed "agents". We always have heterogeneous, open sets (or "collectives") of agents, where "open" means that at any time agents may leave or enter a cooperating crowd. Cooperation of agents may take place on several levels, e.g., perception may be influenced by information from other agents, intention detection may be based on information from other agents or estimated intentions may be fused on an even higher level of abstraction. Intention detection in a distributed way means that basically every agent is able to use all available information or knowledge from other agents on different levels to detect intentions of VRUs or to fuse intention estimates of other agents with own estimates. Knowledge about intentions of VRUs is necessary for anticipatory, automated driving, warning VRUs, etc. Such reactions are not in our focus.

## State of the art:
### VRU perception including cooperative perception

Since the introduction of the HOG (Histogram of Oriented Gradients) descriptor by Dalal and Triggs [4] in 2005, a large number of methods has been developed for image-based detection of pedestrians. In [5] Benenson et al. compared 40+ methods for pedestrian detection. The detectors could roughly be grouped into variants of Deformable Parts Models (DPM), Deep Networks and Decision Forests.

Compared to the work on pedestrians, there is still much less research on the detection of cyclists. Several approaches apply video-based methods for pedestrian detection to cyclists. Li et al. and Hyunggi et al. point out that cyclist detection is more challenging due to the different appearances of a cyclist ( [6] and [7]). Li et al. propose a monocular vision system for vehicle on-board detection of crossing cyclists in an urban traffic scenario using a modified HOG descriptor. Hyunggi et al. use the Integral HOG descriptor together with AdaBoost and construct a cascade of classifiers for different views of the bicycle. In [5] a monocular vision-based 3D tracking approach is proposed to detect bicycles. The object detection uses a mixture of multiscale deformable part models in combination with a latent SVM.

In a non-vision-based approach Wang et al. [9] discuss earlier work on segmentation and classification of cars, pedestrians, bicyclists, and background using data of a 3D laser scanner. The respective algorithms include existence-based tracking using a multiple-model probability hypothesis density filter.

Many dangerous situations occur at intersections due to occluded road users that cannot be detected from a vehicle. Therefore, several projects worldwide, e.g. the European project SAFESPOT [10], subproject INTERSAFE of IP PreVENT, INTERSAFE-2 [11], or the German project Ko-PER of the Ko-FAS research initiative [9] address infrastructure-based perception aiming at an improvement of road safety by combining infrastructure information with local vehicle data. However, the focus in these projects was on establishing C2I- and C2C-communication and fusing information about other, eventually occluded vehicles.

**VRU intention detection**
Intention detection of VRUs is an active research field. It is important for advanced driver assistant systems (ADAS) applications using path prediction for driver warnings or autonomous intervention. Naujoks investigated on the required forecasting time horizon with respect to the time to collision [12] using a driving simulator. The results are, a driver should be informed in best case 2 to 3 seconds, but at least 1 second before a potential collision happens.

The standard technique for tracking and estimation of moving objects are Bayesian recursive state estimator [14] and filters derived from it [14].

Hermes et al. [16] use trajectory classification together with underlying particle filters to predict vehicle trajectories. In [15] Meissner et al. use a multiple-model probability hypothesis density filter to track and predict pedestrians' movements. Due to the pedestrian's ability to quickly change the direction of movement, these models are limited concerning their predictive quality. Approaches aiming to resolve these limitations are from Kooij et. al [18] and Keller [19]. Another infrastructure-sensor trajectory-based tracking approach that uses a video-based descriptor has been introduced by Goldhammer et. al. [20]. It takes the body language into account and uses machine learning techniques, i.e. artificial neural networks, to forecast the pedestrian's trajectory.

A related collaborative VRU intention detection approach, which makes use of organic- and soft computing techniques, is presented in [21].

**Activity Recognition with body-worn**
The use of additional body-worn sensors for intention detection is linked with activity recognition [20], [21] and context-aware computing [24]. An approach to predict intentions of cyclists using smart devices [25] showed promising results concerning the prediction accuracy. They investigated the likelihood of collisions between cyclists and cars, while considering non-constant velocities. A study supporting the long-term goal of protecting cyclists was performed in [26]. The authors developed a system, which incorporates a vehicle with the ability of C2X communication and a cyclist with a WiFi enabled smartphone. In [27], the authors proposed a prototype system using a smart device for a

Car2Pedestrian communication, transmitting the type of movement to an approaching car.

**Cooperative learning and knowledge acquisition**
Distributed intelligent systems that work in a collaborative manner recently became an active research issue (for an overview see, e.g. [96]). Often, information is locally acquired and pre-processed and then sent to special processing units (which can be either centralized or distributed). Also related to our approach, is work dealing with collaborating agents that learn from each other [25]. Other related techniques are from the organic computing domain [26], i.e. collaborative learning in dynamic, distributed environments by means of knowledge exchange. These techniques are based on probabilistic modelling [30] [27], considering inherent uncertainty of knowledge.

# Global Vision

For our approach, we envision the following future traffic scenario including automated driving: Cars, trucks, and other road users equipped with sensors, electronic maps, Internet connection, etc. each determine and continuously maintain a local model of the surrounding traffic situation. This dynamic model results from cooperation with other vehicles in the local environment (e.g. based on ad hoc networks) and it is also based on information from some other traffic participants, such as VRUs – if equipped with appropriate mobile devices, such as smartphones, smartwatches, etc. – and traffic infrastructure – e.g. urban intersections equipped with sensors, such as cameras or laser scanners. Their traffic model arises from a cooperative perception and – projected into the future – a cooperative forecast of traffic situations. Altogether, the models are based on the collective intelligence of the participating vehicles, VRUs, etc. They have to reflect the degree of uncertainty concerning observations, model parameters, and estimates (e.g. concerning future positions of VRUs). Basically, these models, which can be seen as distributed knowledge bases, allow for locally and cooperatively reached decisions such as trajectory planning in automated driving, VRU warning, control of infrastructure (e.g. traffic lights), etc.
We are convinced that pedestrians and cyclists will still play a significant role in the traffic of the future with automated, cooperative vehicles – particularly in urban areas. One key challenge is that vehicles act in such a mixed traffic with VRUs anticipatory, considerate, safe, compliant to traffic rules and, nevertheless, quickly, avoiding abrupt maneuvers. A cyclist, for example, should be passed by a vehicle, if the cyclist keeps going straight on. Cyclists looking back and signaling with the left hand to turn left should not be passed (in right hand traffic). Many similar scenarios in which VRUs signal their future behavior with gestures can be identified. The gesture can even degrade to, for example, standing still in a specific direction at a specific location or eye contacts etc. While connected road users can exchange information via electronic communication units and, therefore, are able to drive coordinated and predictable trajectories, this is not the case with unconnected road users, such as pedestrians and cyclists. The objective and global vision of the proposed approach is that automated, cooperative vehicles take into account VRU behavior. In doing so, it is important that VRU intentions can be detected in a very early phase. Cooperative methods including vehicle-based, infrastructure-based, and mobile device-based perception and intention detection techniques must be developed to cope with occlusions in urban areas and to significantly increase reliability and robustness by resolving implausibilities. Mobile devices carried by VRUs will be applied to detect and transfer intentions to other connected road users (incl. the infrastructure).

## Approach

Our architecture concept is shown in Figure 3. The schematic describes the cooperative system from the point of view of an ego-vehicle and contains multiple, optionally available agents (participants in a cooperation), including infrastructure, body-worn mobile devices, and other vehicles. All agents (including the ego-vehicle) have the ability to perform local VRU detection and intention detection, as well as the ability to interact by exchanging information via wireless communication. Ego-vehicle information and fused information of earlier stages are always available to the ego-vehicle. For the sake of clarity, the corresponding arrows are not shown in Figure 3. Fusion stages of other agents together with their outputs are also not shown in this figure.

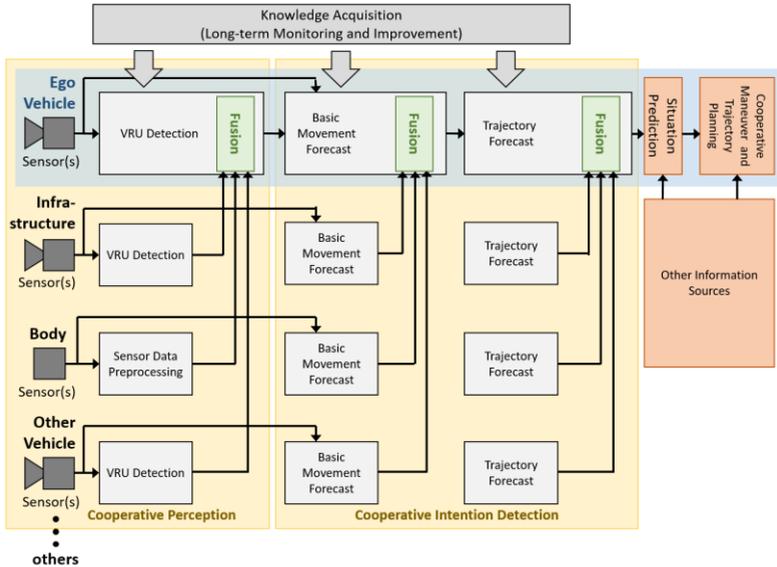

*Figure 3: System architecture for cooperative VRU perception and intention detection.*

Collaboration is possible by exchanging information on different levels. In Figure 3 all levels are shown. Techniques are developed for all levels and subsets based on quality, run-time, latency and data required for communication are investigated. In the stage of cooperative perception, each agent acquires sensor data and uses it for local VRU detection and tracking in its field of view. The local object and track information is then sent to the ego-vehicle and fused to a virtual VRU traffic map. VRU detection results of other agents may additionally control VRU detection of the ego-vehicle. The cooperative intention detection stage consists of basic movement forecast and trajectory forecast. The basic movement forecast is initially performed by each agent separately optionally using locally perceived raw data and the local/fused result of the perception stage. The results can either be features or basic movement forecasts or both. Accordingly, we perform a feature-based or a decision-based fusion (fusion types A and B in Figure 4). The subsequent trajectory forecast stage follows the same strategy: Information may be fused in two different ways, before or after the regression step used to forecast trajectories.

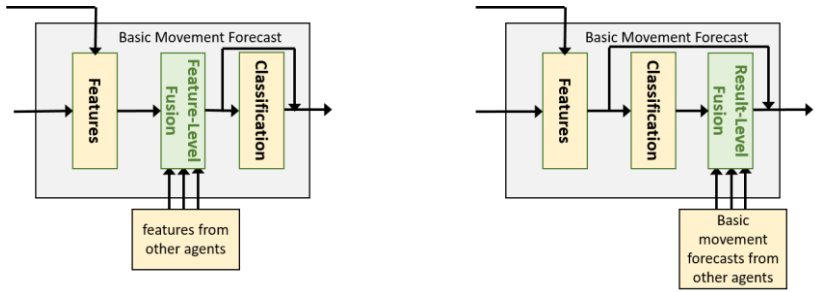

*Figure 4: Feature-based (left) and decision-based (right) fusion for basic movement forecast.*

In the stages of cooperative perception and intention detection, agents may make use of adaptive machine learning algorithms, and permanently expand their existing knowledge by acquired scenarios to improve detection and forecast quality (see knowledge acquisition in Figure 3). The fused results of cooperative intention detection are used in the subsequent stages as basis for situation prediction and cooperative maneuver and trajectory planning. Both steps will get information from other sources, such as, e.g. trajectory forecasts of other vehicles. The approach does not cover these stages, which are labeled with red color in Figure 3.

## VRU Detection and Tracking

The first step of the knowledge acquisition consists of the detection of VRUs. The goal is to develop methods to detect and track VRUs in image sequences provided by cameras mounted in a vehicle or on some infrastructure (e.g. an intersection). Transmitted hypotheses, created by other agents, are used to improve the detection. The results are forwarded to the intention detection. They contain at least information on the position of a VRU, its geometry, and covariances (uncertainty estimates). Algorithms to detect pedestrians are available from Ko-PER, which allows concentrating on the more complicated cyclist detection.

Two main approaches are in the focus of our research. On the one hand, feature descriptors like HOG-Features, as well as multi-scale features for images in combination with soft computing methods, e.g. SVMs or ANNs, are investigated. These techniques have in common, that they are supposed to work on any viewing angle, on any camera resolution, and no matter if they are applied to image

ROIs or to the whole image. They can be easily used and adapted for both, vehicle or infrastructure cameras. On the other hand, a model-based approach is part of the considerations. Different image based techniques to identify geometric primitives are used. As such features are naturally not visible from any side of a VRU, these will not be visible in every frame. Thus, these approaches have to be coupled with other ones.

If the equipped infrastructure is an intersection, it can be assumed, that there have to be multiple stationary cameras to have a proper view at the traffic in all directions. This is referred to as a multi-view setup. Overlapping camera views can be used to confirm hypotheses and to reduce the number of false positives.

The detections made by the feature or model-based approaches are forwarded to a tracker. An IMM (Interacting Multiple Model) filter helps to select an appropriate motion model. Information on the direction of movement provided by the optical flow improves the tracking and could vice versa generate information for the detection. Attention has to be turned on the fact that there may be many cyclists or other moving objects in the scene. On the one hand, each object has to be identified clearly; on the other hand, the filter has to be capable to deal with partly or fully occluded objects.

## Cooperative VRU Intention Detection

The cooperative intention detection stage consists of basic movement forecast and trajectory forecast. Intention detection is initially performed by each agent separately using locally perceived raw data. Resulting local features, predictions or models are then fused and enhanced by information and knowledge from other agents.

**Basic Movement Forecast**

The aim of the basic movement forecasting is to detect and predict movement transitions (e.g. standing to walking) of pedestrians and cyclists at a very early stage. We are interested in movement transitions that indicate the future behavior of VRUs to increase the quality of the subsequent trajectory forecast. Our approach is based on a holistic perception of body language and gestures, e.g. for cyclist we consider the following set of basic behavior: starting, stopping, pedaling, acceleration, deceleration and turning. Gestures, such as, raising the arm to signal a turn are also considered. These basic movement primitives, as well as the gestures are detected using machine learning algorithms, e.g. SVMs, trained on labeled activity data. The model uses the MCHOG-descriptor [32] and other computer vision features extracted from the video. SVM or random forest-based [29] wrapper approaches will be adopted with this objective. A basic movement forecast probability, as well as an accompanying uncertainty of a movement transition is generated. Moreover, not only a joint probability for movement transition is provided, but also a second value describing the uncertainty regarding this probability (cf. the concept of second-order probabilities [30]).

**Trajectory Forecast**
The goal of trajectory forecasting is to provide a short-time trajectory forecast of pedestrians and cyclists, which enables a situation prediction of cooperative, automated vehicles, allowing the vehicles to adapt and plan maneuvers or to take action, such as to perform an emergency brake. Trajectory forecast requires real-time capability and reliable modeling of confidence levels. Our approach is based on supervised machine learning methods, e.g. artificial neural networks (ANNs) and kernel based regression algorithms, e.g. [35], as well as on analytical models [36] for starting, stopping, acceleration, deceleration, and change of direction of bicyclists. We use the result of the basic movement forecast as an important information, e.g. to trigger the analytical models. Regarding the first approach [20] showed that ANNs are capable of dealing with the interesting kinds of occurring pedestrian scenes, such as walking, starting, stopping, or bending, if 3D head trajectories of these movements are used to train the ANN.

This work is a starting point for our supervised machine learning approach to forecast bicycle trajectories. In the second approach we will use analytical models of bicyclist movements to deduce their trajectories. Model parameters of these curves are estimated based on the current and previous states and the forecasted basic movement. The different models can be combined by ensemble approaches [37].

Since neither the machine learning methods, nor the analytical approach, do not per se produce covariance values, such as, recursive Bayesian filters do, confidence models by extended ANN or kernel-based regression technique have to be learned from training data.

## Cooperative Intention Fusion

Combination and fusion takes place on two different levels: feature- and decision-level fusion.

In feature-level fusion, information from different sources is fused to give a complemented and extended view, e.g. Bayesian multi-sensor tracking [38]. A basic assumption of feature-level fusion is that the cooperating agents share at least a part of their feature spaces. The feature-level fusion is based on weighted orthogonal polynomial approximations, allowing to exploit some specific properties of the basis polynomials and the approximation technique. This is at first the availability of an efficient update mechanism, allowing fast incremental approximations [34], and at second the capability to weight, i.e. integrate, information individually. This fast update mechanism allows to update the approximations efficiently at runtime and the weighting allows to decay outdated information (e.g. exponential smoothing), to emphasize on more recent or certain information. The use of polynomial approximations allows to cope with situations where, e.g. due to communication problems, information arrives late or not in the correct temporal order. This makes them perfectly applicable in vehicular ad hoc networks.

In decision-level fusion predictions from different road users, are combined, in a cooperative, e.g. weighted by their certainty, or competitive fashion. The techniques used here are related to ensemble techniques [35], [30] but also to model- or knowledge-fusion techniques, borrowed from the organic computing domain [36]. The approaches follow an archetype of the bagging mechanisms in machine learning and fusion based on second-order probabilities [42].

## Communication Strategy

The cooperative intention fusion approach has to consider communication related problems, such as delays and communication cost, as well as aging of information. Thus, data might be expensive to gather or even outdated. On the other hand, the communication medium is shared and limited, as are the available computational resources. We assume that simple communication strategies will not work (e.g. broadcasting all available information on the one hand or only reacting on requests on the other). Adaptive strategies are needed that depend on the current traffic situation (e.g. number of road users) or prioritize certain information (e.g. pedestrian behind truck is starting to enter the road). Moreover, the communication strategy must also consider the ability to cope with an open set of agents, in a sense any time new participants may enter the system or present participants may leave. Therefore, organic computing techniques, such as novelty detection [38] can be used to detect and quantify the urgency and novelty in addition to their uncertainty in order to prioritize it. In addition, active learning techniques [44] from the domain of autonomous learning, which allow for integration of bandwidth limitation and costs in the communication strategy are applied. In order to quantify these limitations a detailed modeling of the vehicular ad hoc networks (VANets), the communication delays and costs (e.g. probabilistic model) is required [40].

## VRU as Additional Information Sources

Using body-worn sensors and mobile devices (e.g. smartphones and smartwatches) the VRU itself can become an additional source of information, complementing others' sensors, such as video-based techniques in vehicles or infrastructure.

The VRU is then able to help to improve the forecasting by informing vehicles and infrastructure about position, kind of movement (e.g. standing, walking) including forecast of movement transition, and category, such as pedestrian, cyclist, or other. These techniques based on body-worn sensors and mobile devices such as smartphones are much less accurate concerning the absolute positioning than, e.g. video-based techniques, but (1) there are many situations where even rather imprecise information might be very helpful (e.g. if the view from a car to a VRU is occluded by a vehicle and other vehicles are not able to deliver that information), (2) for relative trajectory forecasts and determination of the direction of motion (e.g. away from the curbside) and (3) sensor systems in mobile devices will be continuously improved (e.g. with Galileo, a global navigation satellite system). Furthermore, the recognition and forecasting process can be enriched by additional context information, i.e. the use of additional information sources such as microphones but also calendars, or navigation apps.

## Acknowledgement

This work was funded within the priority program "Cooperatively Interacting Automobiles" of the German Science Foundation DFG.
The author acknowledges the fruitful collaboration with the project partners.